\documentclass[letterpaper, 10 pt, conference]{./support/ieeeconf}
\usepackage{times}
\usepackage{amsmath}
\usepackage[pdftex]{graphicx}
\usepackage{caption}
\usepackage{subcaption}
\usepackage{amsmath,amssymb,amsopn,amstext,amsfonts}
\usepackage{cancel}
\usepackage[space]{cite}
\usepackage{pdfsync}
\usepackage{balance}
\usepackage{color}
\usepackage{algorithm}
\usepackage{algorithmic}
\usepackage{url}
\usepackage{booktabs}
\usepackage{threeparttable}
\usepackage[linkcolor=black,citecolor=black,urlcolor=black,colorlinks=true]{hyperref}
\usepackage{multirow}

\usepackage{graphicx}
\usepackage{epstopdf}
\usepackage{array}
\usepackage{verbatim}
\usepackage{makecell}
\IEEEoverridecommandlockouts

\newcommand{\newrevise}[1]{\textcolor{black}{#1}}

\newcommand{\revise}[1]{\textcolor{black}{#1}}

\graphicspath{{./figure/}}

\DeclareGraphicsExtensions{.pdf,.png,.jpg,.eps,.PNG}
\title{\LARGE \bf BLOS-BEV: Navigation Map Enhanced Lane Segmentation Network, \\ Beyond Line of Sight}
\author{Hang Wu, Zhenghao Zhang, Siyuan Lin, Tong Qin$^*$, Jin Pan, Qiang Zhao, Chunjing Xu, and Ming Yang
		\thanks{
					Hang Wu, Zhenghao Zhang,  Siyuan Lin, Jin Pan, Qiang Zhao and Chunjing Xu are with IAS BU, Huawei Technologies, Shanghai, China.
     Tong Qin and Ming Yang are with the Global Institute of Future Technology, Shanghai Jiao Tong University, Shanghai, China.
		{\tt\small  \{wuhang12, zhangzhenghao6, linsiyuan1, panjin5, zhaoqiang20, xuchunjing\}@huawei.com, \{qintong, mingyang\}@sjtu.edu.cn}. 
		{  $^*$ is the corresponding author}.
	}}

\begin{document}

\maketitle
\thispagestyle{empty}
\pagestyle{empty}

\begin{abstract}

Bird's-eye-view (BEV) representation is crucial for the perception function in autonomous driving tasks. It is difficult to balance the accuracy, efficiency and range of BEV representation.
The existing works are restricted to a limited perception range within $50$ meters. 
Extending the BEV representation range can greatly benefit downstream tasks such as topology reasoning, scene understanding, and planning by offering more comprehensive information and reaction time.
The Standard-Definition (SD) navigation maps can provide a lightweight representation of road structure topology, characterized by ease of acquisition and low maintenance costs.
An intuitive idea is to combine the close-range visual information from onboard cameras with the beyond line-of-sight (BLOS) environmental priors from SD maps to realize expanded perceptual capabilities.
In this paper, we propose BLOS-BEV, a novel BEV segmentation model that incorporates SD maps for accurate beyond line-of-sight perception, up to $200$m. 
Our approach is applicable to common BEV architectures and can achieve excellent results by incorporating information derived from SD maps. 
We explore various feature fusion schemes to effectively integrate the visual BEV representations and semantic features from the SD map, aiming to leverage the complementary information from both sources optimally.
Extensive experiments demonstrate that our approach achieves state-of-the-art performance in BEV segmentation on nuScenes and Argoverse benchmark. 
\newrevise{Through multi-modal inputs, BEV segmentation is significantly enhanced at close ranges below $50m$, while also demonstrating superior performance in long-range scenarios, surpassing other methods by over $20\%$ mIoU at distances ranging from $50\text{-}200m$.}

\end{abstract}

\section{Introduction}
\label{sec:intro}

$\quad$Accurate lane perception is the fundamental function of autonomous vehicles.
However, the irregular and complicated road structure makes it difficult to identify accessible lanes precisely, especially in \newrevise{complex} urban scenarios.
Traditionally, a High-Definition (HD) map is required for autonomous driving in urban scenario, which provides accurate road topological structure.
It is well known that HD map lacks scalability, which limits the wide usage.
Recently, Bird's-eye-view (BEV) networks \cite{li2022bevformer,liang2022bevfusion,yang2023bevformer,zhou2023bev, zou2023diffbev,man2023bev} are widely used for lane perception online.
The BEV perception provides a compact and accurate representation of the surroundings, \revise{providing a flattened top-down perspective essential for path planning and prediction.}

\begin{figure}[t]
	\centering
	\includegraphics[width=0.45\textwidth]{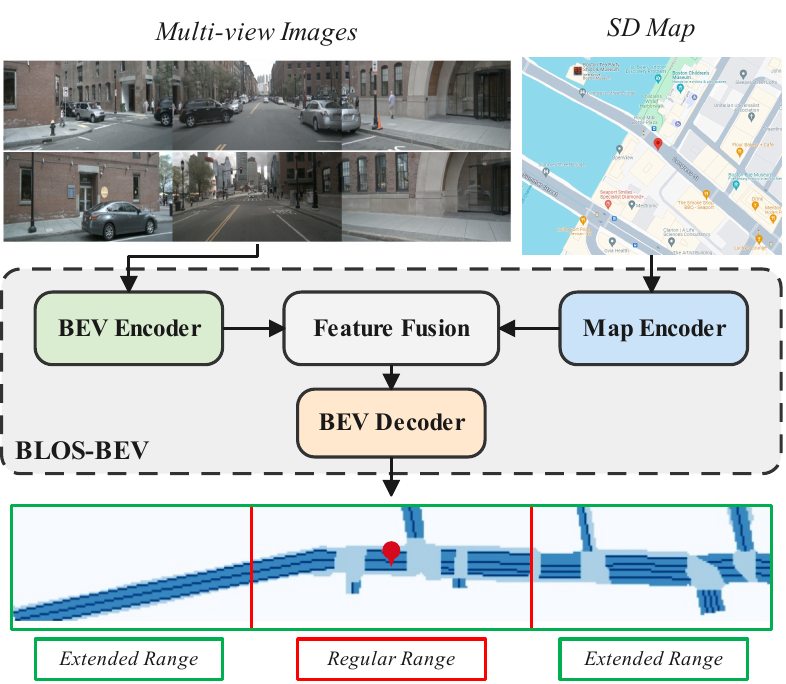}
	\caption{
 The BLOS-BEV architecture. 
 BLOS-BEV effectively integrates the complementary information from surround-view images and SD maps. 
 By fusing visual information and geometrical priors, BLOS-BEV produces BEV semantic segmentation that far exceeds the range of previous methods, 
 enabling extended-range scene parsing critical for safe autonomous driving.
 The video demonstration can be found at: \url{https://youtu.be/dPP0_mCzek4}.
 }
	\label{blos_bev_overview} 
\end{figure}

While the significance of BEV perception is acknowledged, its perceptual range remains relatively unexplored. Moreover, a common limitation observed across existing methods is their BEV range, typically extending up to around $\pm 50$ meters, as seen in previous works such as \cite{roddick2020predicting, LSS, zhou2022cross}.
The restricted range leads to a lack of meaningful contextual understanding at longer distances.
This limitation is primarily attributed to the constrained resolution of onboard cameras and the presence of visual occlusions.
However, there is a critical need to extend the perception range in scenarios that demand a comprehensive understanding of the surroundings, especially in high-speed or long-range planning across large curvature curves.
An expanded environmental perception range correlates with improved autonomous driving safety performance and trajectory smoothness.

In autonomous driving, Standard-Definition (SD) maps serve as lightweight semantic navigation solutions, contrasting with HD maps in terms of detail and resource requirements, making them a readily available alternative.
Although in low accuracy and with blurry elements, SD maps can provide rich semantic information and topological priors, such as road curvature and connectivity, that can be correlated with environmental perception. 
This facilitates localization as well as higher-level contextual understanding of the surroundings. 
Despite these advantages, the fusion of SD maps with learned BEV perception remains unexplored. This untapped potential presents an opportunity to significantly push the frontiers of long-range BEV understanding.

To address the challenge of the perception field, we propose a novel approach, BLOS-BEV, that combines SD map priors with surround-view images, significantly extending the perceptual range of BEV beyond line-of-sight, achieving up to $200$ meters.
This provides downstream tasks, such as prediction and planning, with more operating space significantly.
Our main contributions are summarized as follows:
\begin{itemize}
    \item We propose a novel network to fuse navigation maps with multi-view images for long-range BEV segmentation. Besides, our architecture demonstrates universality, seamlessly integrating into existing BEV methods.
    \item \revise{We investigated various feature fusion methods to combine visual BEV features with semantic features from navigation maps, aiming to derive an optimal representation that effectively captures connections between these two complementary sources of information.}
    \item Our model achieves state-of-the-art BEV segmentation performance \revise{at both short and long distances, attaining  beyond line-of-sight perception capabilities. This advancement establishes a solid foundation for enhanced safety in autonomous driving.}
\end{itemize}  

\section{literature review}
\label{sec:realted_work}

\subsection{BEV Segmentation}

Efficient and accurate BEV segmentation is a critical task in autonomous driving, enabling the understanding of the ego-vehicle's surroundings and supporting downstream tasks such as behavior prediction and planning.
Recent works \cite{wang2019monocular, Reiher_2020, LSS, roddick2020predicting, zhou2022cross, BEVDepth, gosala2022bird, xie2022m, ng2020bev, zhao2024improving} typically involve a transformation process that converts perspective images from cameras or sensors into a top-down, BEV representation, followed by predicting semantic labels for each pixel, such as drivable areas, lanes, obstacles, and more.
CAM2BEV\cite{Reiher_2020} corrects the flatness assumption error and occlusion issues encountered with Inverse Perspective Mapping (IPM)\cite{mallot1991inverse} projection.
LSS\cite{LSS} generates a frustum point cloud to implicitly estimate depth distribution, then projects the frustum onto the BEV plane using camera intrinsics and extrinsics for semantic segmentation.
BEVDepth\cite{BEVDepth} leverages explicit depth supervision to optimize depth estimation, boosting performance.
CVT\cite{zhou2022cross} uses a camera-aware attention mechanism to learn a mapping from Perspective View (PV) to a BEV representation, without explicit geometric modeling.
BEVSegFormer\cite{Peng_2023} uses a novel multi-camera deformable cross-attention module that eliminates the need for camera intrinsic and extrinsic parameters.
DiffBEV\cite{zou2023diffbev} and DDP\cite{Ji_2023} have explored using diffusion models to progressively refine BEV features, yielding more detailed and enriched feature representations. 

\subsection{SD Map for Autonomous Driving}
SD navigation maps provide spatial information for various navigation systems, such as autonomous driving. 
Information includes road geometry, topology, attributes, and landmarks. 
Recently, more research has focused on how to use SD maps in autonomous driving tasks. 
Panphattarasap \textit{et al.} \cite{automated_map_reading} introduced a novel approach to image$\text{-}$based localization in urban environments using semantic matching between images and a $2$D map. 
The approach employed a network to detect the features in images and match descriptors from image features with descriptors derived from the $2$D map. 
Zhou \textit{et al.} \cite{image_based_geolocalization} presented a $2.5$D map$\text{-}$based cross$\text{-}$view localization method that fused the $2$D image features and $2.5$D maps to increase the distinctiveness of location embeddings. 
OrienterNet \cite{sarlin2023orienternet} proposed a deep neural network that estimated the pose of a query image by matching a neural BEV with available maps from OSM and achieved meter$\text{-}$level localization accuracy. 

\subsection{\newrevise{Fusing Prior Information} for Road Structure Cognition}

There are some existing works\cite{ben2022toponet, liao2023maptrv2, liao2022maptr, qiao2023end, liu2023vectormapnet} on road structure recognition, however, they solely rely on onboard sensors, which experience degraded performance over extended perception ranges.
Recent research have been focused on exploiting prior information fusion to enhance the resilience and efficiency of online map generation. In the case of NMP \cite{xiong2023neural}, a neural representation is acquired, and a global map prior is constructed from previous traversals to enhance online map prediction. Similarly, in \cite{can2023prior}, optimization in the latent space is employed to acquire a globally consistent prior for maps. \cite{gao2023complementing} addresses the challenge of long$\text{-}$range perception in HD maps by augmenting the images from the camera on board with satellite images. 
Our study shares close affinity with these methodologies; nevertheless, we employ a distinct prior SD maps, characterized by significantly superior compactness in both storage and representation. Additionally, SD maps are widely accessible, contributing to their practicality and ease of utilization in comparison to the approaches mentioned.


\section{methodology} 
\label{sec:methodology}
\subsection{Overview}

\begin{figure*}[t]
	\centering
	\includegraphics[width=0.9\linewidth]{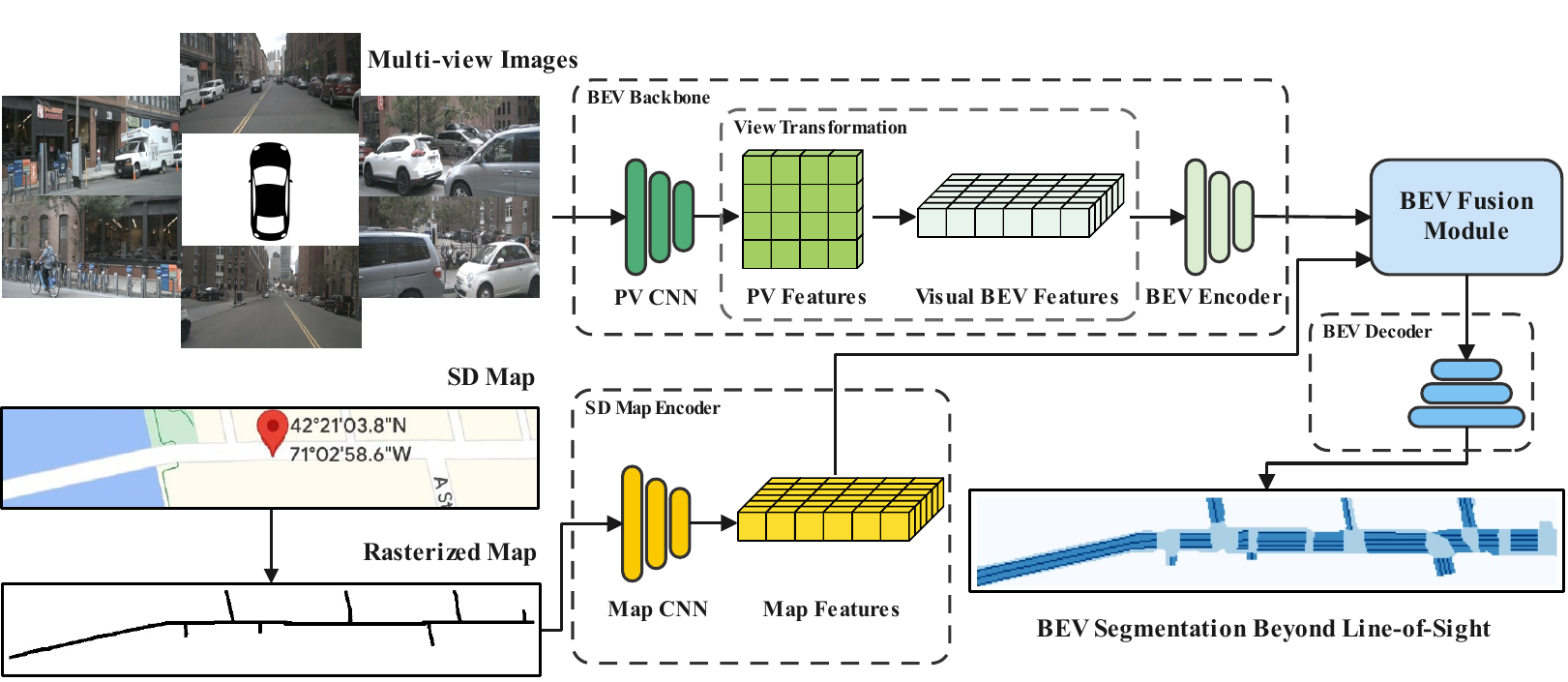}
	\caption{Pipeline of the BLOS-BEV model. The surround-view camera images from the ego vehicle along with a rasterized SD map are fed as inputs. The SD map provides the key road topology. BLOS-BEV effectively fuses the visual features and map encodings through a BEV fusion module. By integrating complementary information from images and maps, BLOS-BEV produces beyond line-of-sight BEV segmentation that substantially exceeds the range of previous methods.}
	\label{fig:pipeline}
\end{figure*}

Our BLOS-BEV framework consists of four main components: the BEV \newrevise{Backbone}, the SD Map Encoder, the BEV Fusion Module, and the BEV Decoder, as shown in Fig.$\;$\ref{fig:pipeline}.
This architecture ultimately enables enhanced perceptual range and planning foresight by synergistically integrating complementary input modalities.

\subsection{BEV \newrevise{Backbone}}
We adopt Lift-Splat-Shoot (LSS)\cite{LSS} as a BEV feature extractor \revise{baseline} due to its lightweight, efficient and easy-to-plug characteristics.
\revise{Other BEV architectures (e.g., HDMapNet \cite{li2022hdmapnet}) are also adaptable within our framework.}
LSS learned the depth distribution of each pixel and used camera parameters to transform the frustum into a BEV representation. 
\revise{The onboard cameras in six orientations (Front, Front-Left, Front-Right, Rear, Rear-Left, Rear-Right) provide the model with a surround-view visual input for comprehensive situational awareness.}
The output \newrevise{of the View Transformation} is the visual BEV feature $F_{v} \in \mathbb{R}^{{H}\times{W}\times{C}}$, \newrevise{where $H \times W$ and $C$ are the resolution and embedding dimension of BEV representation.}
\newrevise{Subsequently, we adapt a 4-stage FPN \cite{lin2017feature} as the BEV Encoder to further encode the BEV features, with each stage halving the height and width while doubling the channel dimension of the feature maps. We select the 2nd stage feature $F_{v_2} \in \mathbb{R}^{{\frac{H}{2}}\times{\frac{W}{2}}\times{2C}}$ and the 4th stage feature $F_{v_4} \in \mathbb{R}^{{\frac{H}{8}}\times{\frac{W}{8}}\times{8C}}$ as the inputs for the BEV Fusion Module.}

\subsection{SD Map Encoder}

\revise{The SD Map Encoder primarily builds upon a convolutional neural network (CNN) architecture, taking the SD map centered at the ego-vehicle's location as input.}

\begin{figure}
\centering
\begin{subfigure}{.48\textwidth}
    \centering
    \includegraphics[width=0.96\linewidth]{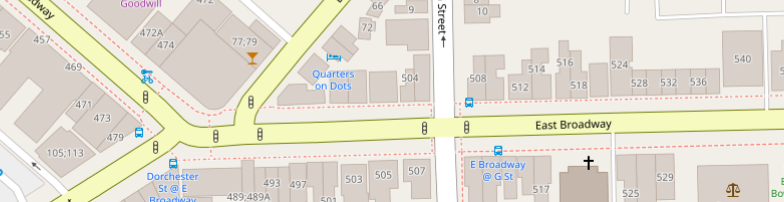}\hspace*{1cm}
    \caption{Visualization of original OpenStreetMaps.}
    \label{sd_osm}
\end{subfigure}
\begin{subfigure}{.48\textwidth}
    \centering
    \includegraphics[width=0.96\linewidth]{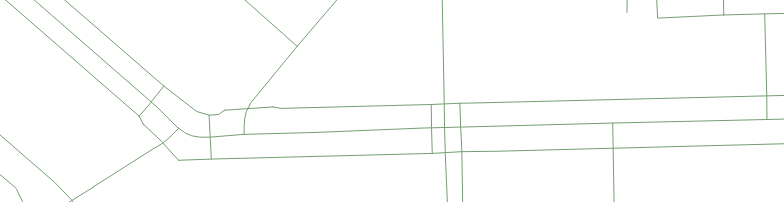}\hspace*{1cm}  
    \caption{Rasterized OpenStreetMaps.}
    \label{sd_raster}
\end{subfigure}

\caption{Comparison of original and rasterized SD maps. The rasterization retains only the key road layout, reducing clutter while providing the essential environmental context for BEV scene understanding. This demonstrates our map preprocessing and rasterization approach to generate a clean topological representation as input to SD Map Encoder.}
\label{fig:sd_sd}
\end{figure}

\noindent
\textbf{Map Data:} 
We leverage the \newrevise{OpenStreetMap (OSM) \cite{OpenStreetMap}}, a crowd-sourced project that provides free and editable maps of the world, to provide prior road information. 
OSM contains rich information about various geographic features, such as roads, traffic signs, building areas, etc. 
Fig.$\;$\ref{fig:sd_sd}\revise{(a)} \revise{illustrates a typical representation of an OSM.}

\noindent
\textbf{Pre-Processing:} 
\revise{To simplify SD map data and eliminate the impact of irrelevant map elements on the final task, we rasterized only the road skeleton from OSM. This enables the SD Map Encoder to focus more precisely on the topological structure of the roads. Fig.$\;$\ref{fig:sd_sd}(b) illustrates the result of rasterizing OSM in our approach.}

\noindent
\textbf{Encoding: } 
\revise{Drawing inspiration from OrienterNet\cite{sarlin2023orienternet}, we adopt a VGG\cite{simonyan2014very} architecture as the backbone for our SD Map Encoder. This generates a spatial encoded map representation $F_{sd}$ which preserves the semantic, positional, and relational information offered by the prior OSM environmental annotation. \newrevise{To align the sizes of BEV features for fusion, we selected the SD map features $F_{sd_2} \in \mathbb{R}^{{\frac{H}{2}}\times{\frac{W}{2}}\times{2C}}$ and $F_{sd_4} \in \mathbb{R}^{{\frac{H}{8}}\times{\frac{W}{8}}\times{8C}}$ from the corresponding stages of the SD Map Encoder as the inputs for the BEV Fusion Module.}}


\begin{figure}[t]
	\centering
	\includegraphics[width=0.45\textwidth]{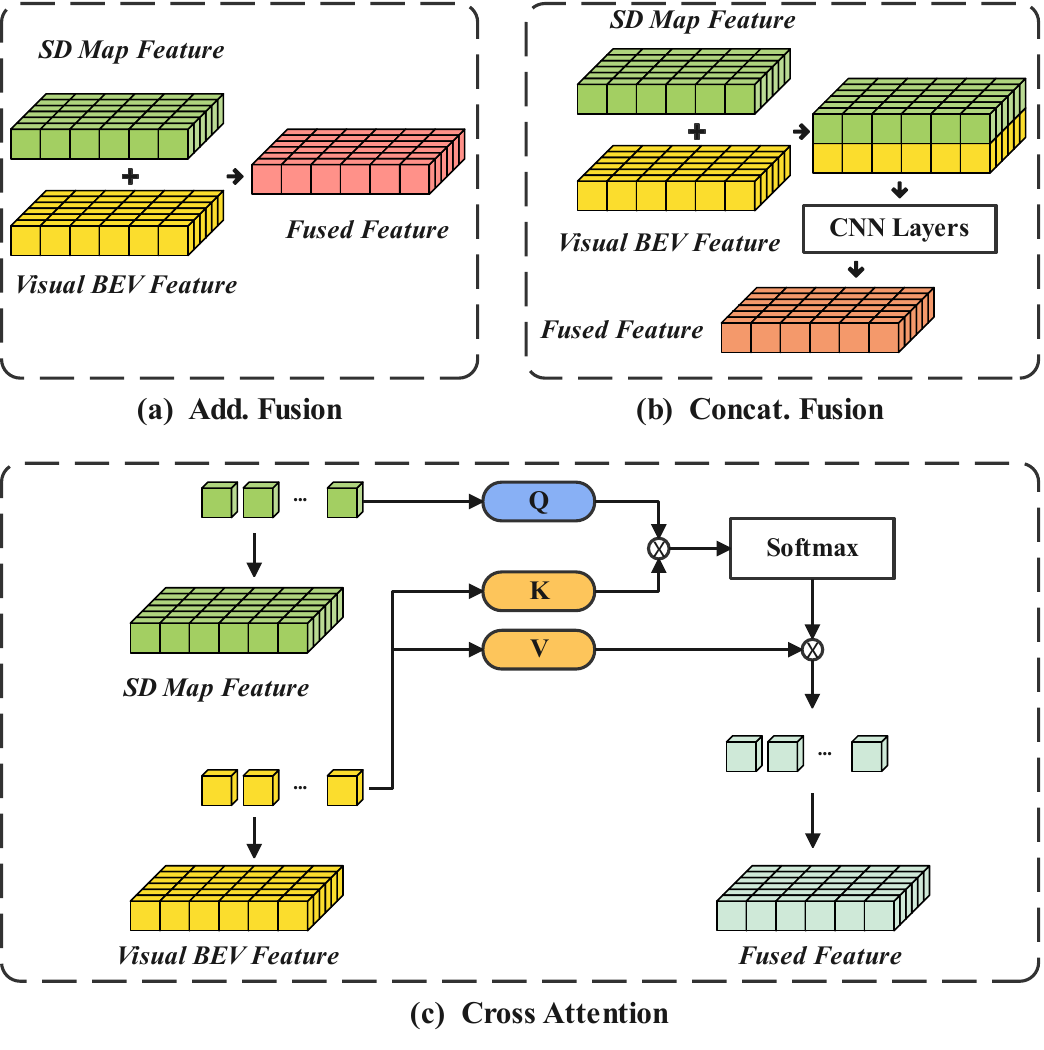}
	\caption{Alternative techniques explored for fusing BEV features and SD map representations in BLOS-BEV. (a) Element-wise addition of BEV and map encodings. (b) Concatenation of BEV and map features along channel dimension, followed by $3\times3$ convolutions to reduce channels. (c) Cross-attention mechanism where map encodings query visual BEV features.}
	\label{fig:sd_fusion}
\end{figure}

\subsection{BEV Fusion Module}
A key contribution of BLOS-BEV is exploring different fusion schemes to combine the visual BEV features and SD map semantics for optimal representation and performance. Three prevalent approaches are addition, concatenation, and cross-attention mechanism. 
Our experiments assess these strategies to determine the most effective yet efficient integration technique for enhanced navigational foresight. 

\newrevise{Since both the BEV branch and the SD map branch provide high and low resolution features of different sizes, we apply the same fusion operation to features of the same size from both branches, resulting in two multi-modal fusion features, $F^h_{fuse}$ and $F^l_{fuse}$, with high and low resolution, respectively. To simplify notation, we use $F_{v}$ and $F_{sd}$ to denote high or low-resolution BEV features ($F_{v_2}$ or $F_{v_4}$) and SD map features ($F_{sd_2}$ or $F_{sd_4}$), respectively.
}
Similarly, we denote $F^h_{fuse}$ and $F^l_{fuse}$ collectively as $F_{fuse}$.

\noindent \textbf{Element-wise Addition:} 
Since the visual BEV features $F_{v}$ and \newrevise{SD map} features $F_{sd}$ are in the same shape, we fuse them \newrevise{via element-wise addition} (see Fig.$\;$\ref{fig:sd_fusion}(a)).
\newrevise{The fused feature $F_{fuse}$ is computed as follows:}
\begin{align}
F_{fuse} = F_{v} + F_{sd}
\end{align}
\noindent \textbf{Channel-wise Concatenation:} We also explore concatenating the BEV and map representations along the channel dimension, using $2$ convolutional layers with $3\times3$ kernels to integrate the concatenated features and reduce channels (see Fig.$\;$\ref{fig:sd_fusion}(b)). 
\newrevise{The fused feature $F_{fuse}$ obtained through concatenation is computed as follows:}
\begin{align}
F_{fuse} = Conv_{3 \times 3}(Concat(F_{v}, F_{sd}))
\end{align}

\noindent
\textbf{Cross-Attention Mechanism:} Furthermore, 
\newrevise{we employ a cross-attention\cite{ashish2017attention} mechanism to fuse the SD map features with the visual BEV features. 
Cross-attention applies inter-modal gating to selectively emphasize the most relevant features from each encoder per spatial location.
Specifically, we use $F_{sd}$ as the Queries $\mathbf{Q}$ and $F_{v}$ as the Keys $\mathbf{K}$ and Values $\mathbf{V}$(see Fig.$\;$\ref{fig:sd_fusion}c).
Our motivation for this design is that since $F_{sd}$ encodes prior information beyond the perception range, querying the local visual features $F_{v}$ allows better reasoning about road structures outside the field of view}.
\newrevise{The fused feature $F_{fuse}$ obtained through cross-attention is computed as follows:}
\begin{align}
F_{fuse} &= Attn \, Block(F_{sd}, F_{v}, F_{v}) \\
Attn \, Block(Q, K, V) &= Attn(QW_i^Q, KW_i^K, VW_i^V) \\
Attn(Q, K, V) &= softmax(\frac{QK^T}{\sqrt{d_k}})V
\end{align}
\newrevise{where $W_i^Q, W_i^K, W_i^V$ are the projection matrices for the $\mathbf{Q}$, $\mathbf{K}$, and $\mathbf{V}$ at the i-th layer, respectively, and $d_k$ is the channel dimension of the feature $\mathbf{Q}$ and $\mathbf{K}$.}

\subsection{BEV Decoder and Training Loss}
In the BEV Decoder, we receive the high and low resolution fusion features, $F^h_{fuse}$ and $F^l_{fuse}$.
We first upsample $F^l_{fuse}$ by a factor of 4, aligning its feature height and width with $F^h_{fuse}$. 
Then we concatenate it with $F^h_{fuse}$ along the channel dimension, followed by two convolutional layers and upsampling, to decode them into a BEV segmentation map of size ${H}\times{W}\times{N}$, where $N$ is the semantic category number.

\newrevise{In the training phase, we use binary cross-entropy (BCE) loss for the category set $\Omega$ that contains the lane, the road, the lane divider, and the road divider:}

\begin{align}
\mathcal{L}_{seg} = - \frac{1}{N}\sum_{c \in \Omega}y_clog(x_c) + (1-y_c)log(1-x_c)
\end{align}
\newrevise{where $x_c$ and $y_c$ are the pixel-wise semantic prediction and the ground-truth label.}


\section{Experiments}
\label{sec:Experiments}

\begin{figure}[t]
	\centering
	\includegraphics[width=0.4\textwidth]{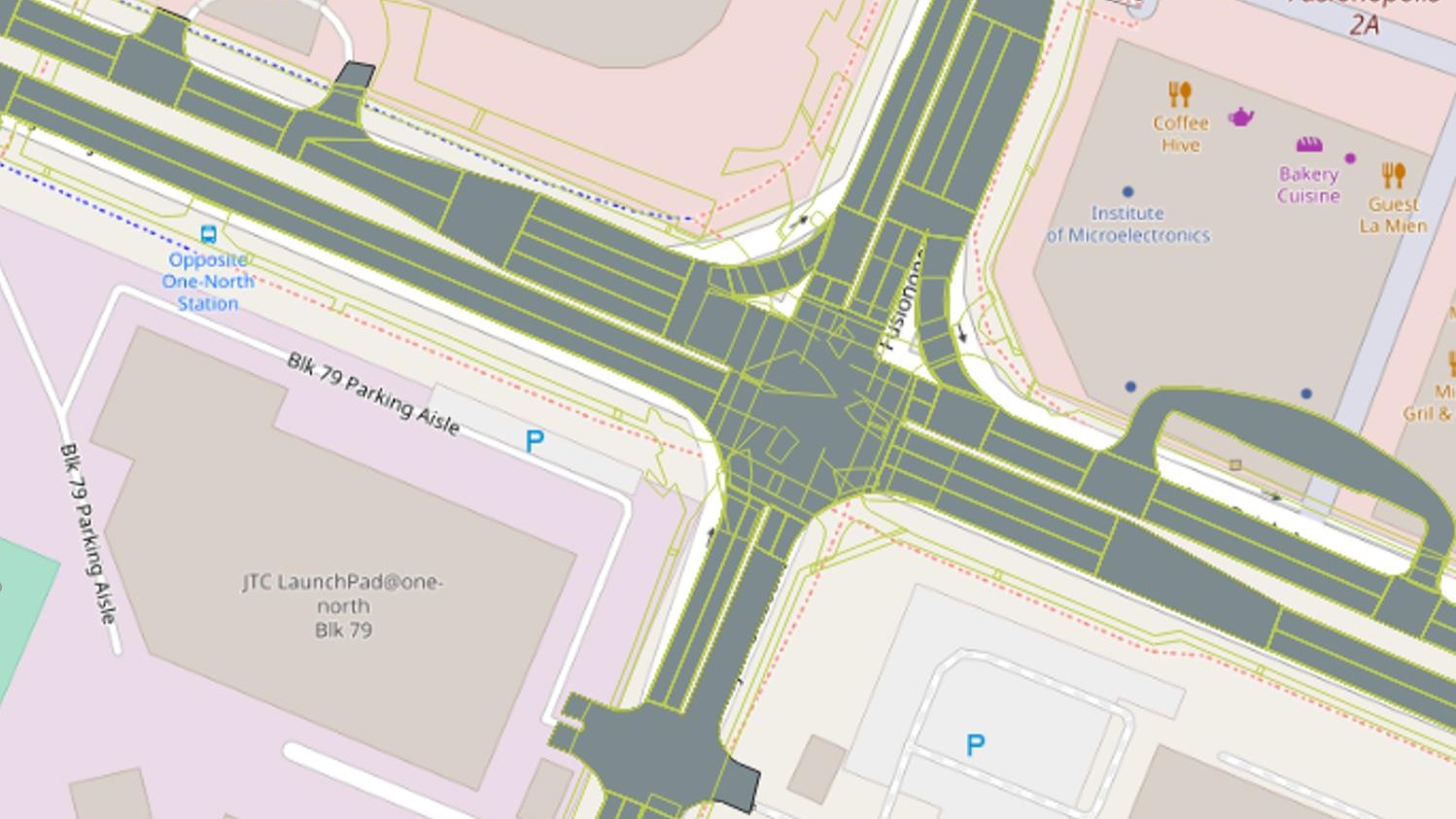}
	\caption{Projection of a nuScenes data onto aligned SD map coordinates, visualized for a local area. The lane and road segment annotations from one nuScenes sequence are transformed and visualized on the SD map.}
	\label{alignment}
\end{figure}

\subsection{Datasets}

\begin{table*}[htbp]
  \centering
  \resizebox{\textwidth}{20mm}{
  \begin{tabular}{c|c|ccc|cccc|cc}
    \toprule
    \multirow{4.0}{*}{\centering Methods}
    &\multicolumn{9}{c}{\centering Eval IoU} \\
    \cmidrule(r){2-11}
    &\multicolumn{1}{c|}{\centering Regular}
    &\multicolumn{3}{c|}{\centering Long Range}
    &\multicolumn{4}{c|}{\centering Overall} 
    &\multicolumn{1}{c}{\multirow{2.5}{*}{\centering Mean}} 
    &\multicolumn{1}{c}{\multirow{2.5}{*}{\centering \revise{FPS}}} \\
    \cmidrule(r){2-9}
    &\multicolumn{1}{c|}{\centering $0\sim50$}
	&\multicolumn{1}{c}{\centering $50\sim100$}
    &\multicolumn{1}{c}{\centering $100\sim150$}
    &\multicolumn{1}{c|}{\centering $150\sim200$}
    &\multicolumn{1}{c}{\centering $lane$}
	&\multicolumn{1}{c}{\centering $road$}
    &\multicolumn{1}{c}{\centering $lane \, divider$}
    &\multicolumn{1}{c|}{\centering $road \, divider$} \\
    \midrule
    HDMapNet\cite{li2022hdmapnet} & $65.56\%$ & $56.06\%$ & $\underline{52.07\%}$ & $47.02\%$ & $65.07\%$ & $51.22\%$ & $14.84\%$ & $17.80\%$ & $56.16\%$ & $12.5$ \\
    PON\cite{roddick2020predicting} & $63.78\%$ & $51.39\%$ & $46.83\%$ & $43.54\%$ & $58.06\%$ & $45.38\%$ & $14.23\%$ & $17.56\%$ & $51.92\%$ & $\underline{15.1}$ \\
    CVT\cite{zhou2022cross} & $\underline{69.12\%}$ & $\underline{57.34\%}$ & $51.32\%$ & $\underline{47.56\%}$ & $\underline{66.23\%}$ & $\underline{52.37\%}$ & $\underline{15.70\%}$ & $\underline{18.79\%}$ & $\underline{57.38\%}$ & $\textbf{16.5}$ \\
    LSS\cite{LSS} & $67.06\%$ & $56.57\%$ & $50.88\%$ & $47.15\%$ & $65.19\%$ & $51.66\%$ & $15.60\%$ & $18.55\%$ & $56.41\%$ & $14.3$ \\
    \midrule
    \revise{BLOS-BEV$\dagger$} & $77.50\%$ & $76.46\%$ & $74.73\%$ & $67.84\%$ & $82.11\%$ & $72.50\%$ & $31.06\%$ & $40.17\%$ & $74.58\%$ & $10.3$ \\
    BLOS-BEV* & $\textbf{79.53\%}$ & $\textbf{77.95\%}$ & $\textbf{76.00\%}$ & $\textbf{70.04\%}$ & $\textbf{83.29\%}$ & $\textbf{74.70\%}$ & $\textbf{34.46\%}$ & $\textbf{42.60\%}$ & $\textbf{76.49\%}$ & $12.8$ \\
    \bottomrule
  \end{tabular}}
  \caption{Performance comparison of Beyond Line-Of-Sight segmentation on nuScenes dataset \cite{caesar2020nuscenes}. We compared our approach (BLOS-BEV$\dagger$ adopts HDMapNet method, BLOS-BEV* adopts LSS method with \newrevise{concatenate} fusion) with previous SOTA methods by dividing the view distance into intervals of 50 meters and covering four major road structure elements. 
  }
  \label{tab:comparation with prior sota}
\end{table*}

We leveraged two autonomous driving datasets $\text{-}$ nuScenes\cite{caesar2020nuscenes} and Argoverse\cite{Argoverse} for the training and validation of our proposed approach.
The nuScenes contains $1000$ driving sequences captured in Boston and Singapore, with a total of $40k$ keyframes. 
Argoverse dataset contains $113$ scenes captured in Miami and Pittsburgh. 
We use the default training/validation split of nuScenes and Argoverse. 

Due to the absence of original SD map data in the nuScenes and Argoverse datasets, we supplemented our dataset by obtaining SD maps from OSM for the corresponding regions.
The nuScenes dataset exhibits misalignment between the HD and SD maps, while the Argoverse dataset does not present these issues.
Through experimentation, we determined a multi$\text{-}$step coordinate transformation was essential to resolve the mismatches. First converting the map coordinate system from EPSG:$4326$ to EPSG:$3857$, applying the map origin offset, and finally transforming back to EPSG:$4326$ yielded correct aligned latitude and longitude coordinates, rectifying the Singapore locale. 
However, in the Boston area, besides coordinate system transformation, an additional $1.35\times$ scaling of the map origin coordinates was required to achieve proper alignment.

With the corrected latitude and longitude coordinates \newrevise{obtained through the aforementioned steps}, we can effortlessly obtain accurate SD map data. 
As shown in Fig. \ref{alignment}, we transform the lane dividers, road segments, and road dividers from one nuScenes scene to the SD map coordinate frame and visualize them projected onto the SD map for validation.

\subsection{Implementation Details}

The training of our model adopts the Adam\cite{kingma2014adam} optimizer with a weight decay rate of \newrevise{$1 e-7$} and an initial learning rate of \newrevise{$1e-3$}. 
All of the experiments used one Nvidia V$100$ GPU and kept the training steps to $200k$ steps. 
The surround-view images are uniformly resized to $352 \times 128$ and then used as input to the model.

BLOS-BEV is built on the top of LSS \cite{LSS}, which is lightweight and fast-inference. 
In the SD Map Encoder module, VGG-$16$ \cite{simonyan2014very} was selected as the encoder. 
The BEV segmentation output range is set to $400 m\times96 m$, with a resolution of $1ppm$ (pixel-per-meter). 
Empirically, we treat $0\sim50m$  as the regular visual range for nuScenes and Argoverse dataset, and $50\sim 200m$ as the long visual range.

\begin{figure*}[t]
	\centering
	\includegraphics[width=0.9\textwidth]{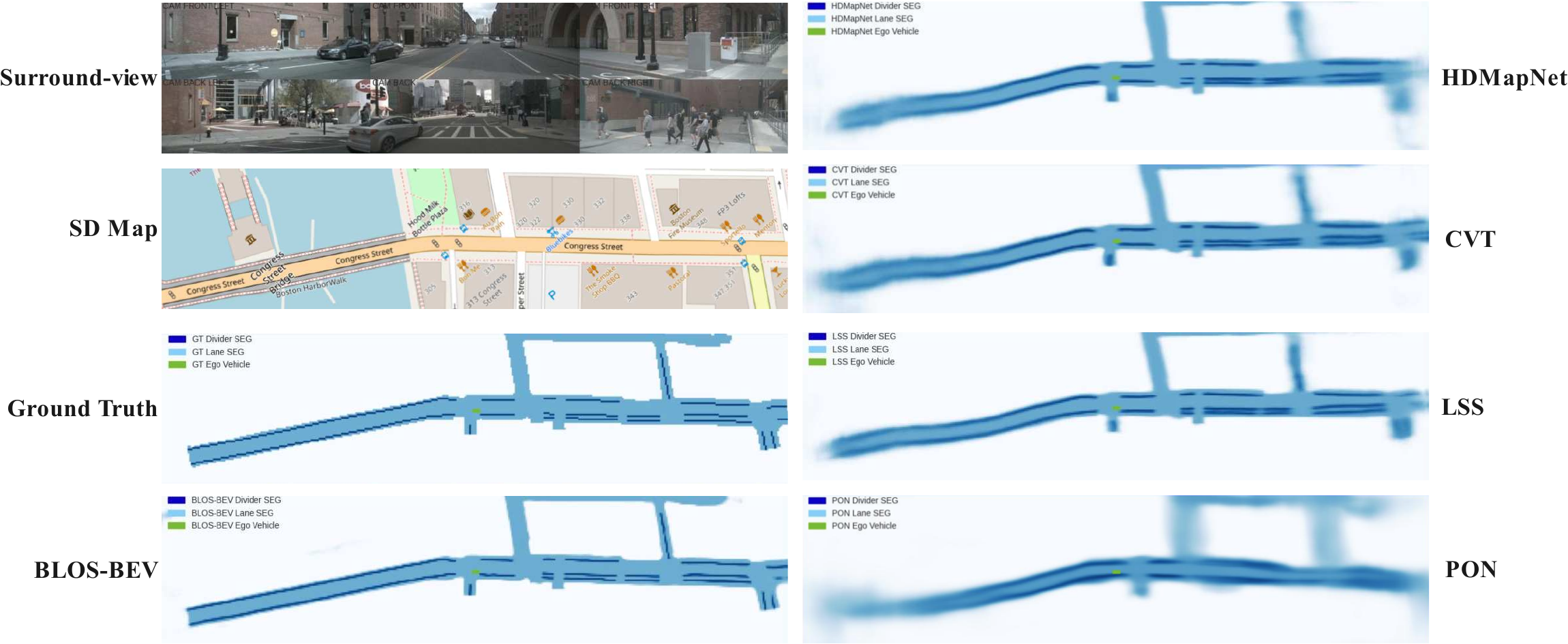}
	\caption{Qualitative comparison of BLOS$\text{-}$BEV against other methods on the nuScenes dataset. The first
 \revise{column of images showcases the surrounding view of the vehicle, the SD map of the current location, BEV segmentation ground truth, and the results of our model BLOS-BEV. For comparison, the second column presents the output results of HDMapNet, CVT, LSS, and PON, models that lack prior information from SD maps.}
 }
	\label{BEV comparison}
\end{figure*}

\subsection{BEV Semantic Segmentation Results}

\begin{table*}[h]
  \centering
  \resizebox{\textwidth}{19mm}{
  \begin{tabular}{c|c|ccc|cccc|c} 
    \toprule
    \multirow{3.8}{*}{\centering Methods}
    &\multicolumn{9}{c}{\centering Eval IoU} \\
    \cmidrule(r){2-10}
    &\multicolumn{1}{c|}{\centering Regular}
    &\multicolumn{3}{c|}{\centering Long Range}
    &\multicolumn{4}{c|}{\centering Overall} 
    &\multicolumn{1}{c}{\multirow{2.3}{*}{\centering Mean}} \\
    \cmidrule(r){2-9}
    &\multicolumn{1}{c|}{\centering $0\sim50$}
	&\multicolumn{1}{c}{\centering $50\sim100$}
    &\multicolumn{1}{c}{\centering $100\sim150$}
    &\multicolumn{1}{c|}{\centering $150\sim200$}
    &\multicolumn{1}{c}{\centering $lane$}
	&\multicolumn{1}{c}{\centering $road$}
    &\multicolumn{1}{c}{\centering $lane \, divider$}
    &\multicolumn{1}{c|}{\centering $road \, divider$} \\
    \midrule
    LSS\cite{LSS} & $67.06\%$ & $56.57\%$ & $50.88\%$ & $47.15\%$ & $65.19\%$ & $51.66\%$ & $15.60\%$ & $18.55\%$ & $56.41\%$ \\
    \revise{Only SD Input} & $62.78\%$ & $63.34\%$ & $62.92\%$ & $61.54\%$ & $75.53\%$ & $66.89\%$ & $12.34\%$ & $13.83\%$ & $62.61\%$ \\
    \midrule
    BLOS-BEV Add. & $78.18\%$ & $76.70\%$ & $74.85\%$ & $68.00\%$ & $81.99\%$ & $73.03\%$ & $32.46\%$ & $41.22\%$ & $75.08\%$ \\
    BLOS-BEV \newrevise{Concat.} & $\textbf{79.53\%}$ & $\textbf{77.95\%}$ & $\textbf{76.00\%}$ & $70.04\%$ & $\textbf{83.29\%}$ & $\textbf{74.70\%}$ & $\textbf{34.46\%}$ & $\textbf{42.60\%}$ & $\textbf{76.49\%}$ \\
    BLOS-BEV Cross-Att. & $77.87\%$ & $77.32\%$ & $75.48\%$ & $\textbf{70.08\%}$ & $83.05\%$ & $74.21\%$ & $30.46\%$ & $37.04\%$ & $75.70\%$ \\
    \bottomrule
  \end{tabular}}
  \caption{Performance comparison of various fuse methods on nuScenes\cite{caesar2020nuscenes} dataset. We test the segmentation performance of various fusion methods under the beyond line-of-sight setting while maintaining the same training epoch number. 
  }
  \label{tab:exploration on fusion method of sd map}
\end{table*}

\begin{table*}[h]
  \centering
  \begin{tabular}{c|c|ccc|cc|c} 
    \toprule
    \multirow{3.8}{*}{\centering Methods}
    &\multicolumn{7}{c}{\centering Eval IoU} \\
    \cmidrule(r){2-8}
    &\multicolumn{1}{c|}{\centering Regular}
    &\multicolumn{3}{c|}{\centering Long Range}
    &\multicolumn{2}{c|}{\centering Overall} 
    &\multicolumn{1}{c}{\multirow{2.3}{*}{\centering Mean}} \\
    \cmidrule(r){2-7}
    &\multicolumn{1}{c|}{\centering $0\sim50$}
	&\multicolumn{1}{c}{\centering $50\sim100$}
    &\multicolumn{1}{c}{\centering $100\sim150$}
    &\multicolumn{1}{c|}{\centering $150\sim200$}
    &\multicolumn{1}{c}{\centering $drivable$ $area$}
	&\multicolumn{1}{c|}{\centering $road$ $boundary$} \\
    \midrule
    LSS(baseline) & $51.4\%$ & $44.5\%$ & $38.5\%$  & $34.8\%$ & $51.3\%$ & $27.7\%$ & $39.5\%$ \\
    \midrule
    BLOS-BEV Add. & $65.1\%$ & $63.3\%$ & $61.4\%$   & $56.9\%$  & $74.6\%$ & $44.1\%$ & $59.5\%$  \\
    BLOS-BEV \newrevise{Concat.} & $66.4\%$ & $64.1\%$ & $61.8\%$   & $57.4\%$ & $74.8\%$ & $45.1\%$ & $60.4\%$  \\
    BLOS-BEV Cross-Att. & $\textbf{67.5}\%$ & $\textbf{66.7}\%$ & $\textbf{64.9}\%$   & $\textbf{60.8}\%$ &$\textbf{78.5\%}$ & $\textbf{46.5}\%$ & $\textbf{62.5\%}$ \\
    \bottomrule
  \end{tabular}
  \caption{We investigate the performance of BLOS-BEV generation on the Argoverse\cite{chang2019argoverse} dataset. The above results show that the fusion of BEV and SD Map features achieves superior performance on both the regular range ($0 \text{-}50 m$) and BLOS range ($150 \text{-}200 m$). In particular, BLOS$\text{-}$BEV Cross$\text{-}$Att achieves the highest mIoU in all ranges.}
  \label{tab:Argoverse_BLOS}
\end{table*}

\begin{table*}[h]
  \centering
  \begin{tabular}{c|c|cccc|c} 
    \toprule
    \multirow{2.3}{*}{\centering Methods}
    &\multicolumn{5}{c}{\centering Eval IoU} \\
    \cmidrule(r){2-7}
    &\multicolumn{1}{c|}{\centering $Train \, Aug$}
    &\multicolumn{1}{c}{\centering $0\sim50$}
	&\multicolumn{1}{c}{\centering $50\sim100$}
    &\multicolumn{1}{c}{\centering $100\sim150$}
    &\multicolumn{1}{c|}{\centering $150\sim200$}
    &\multicolumn{1}{c}{\centering Mean} \\
    \midrule
    LSS(baseline) & $-$ & $67.06\%$ & $56.57\%$ & $50.88\%$ & $47.15\%$ & $56.41\%$ \\
    \midrule
    BLOS-BEV Add. & $\times$ & $42.60\%$ & $40.26\%$ & $39.77\%$ & $37.66\%$ & $40.38\%$ \\
    BLOS-BEV \newrevise{Concat.} & $\times$ & $42.63\%$ & $40.29\%$ & $39.91\%$ & $38.04\%$ & $40.50\%$ \\
    BLOS-BEV Cross-Att. & $\times$ & $43.48\%$ & $42.35\%$ & $41.93\%$ & $39.78\%$ & $42.13\%$ \\
    \midrule
    BLOS-BEV Add. & $\checkmark$ & $69.58\%$ & $61.61\%$ & $58.05\%$ & $53.06\%$ & $61.42\%$ \\
    BLOS-BEV \newrevise{Concat.} & $\checkmark$ & $\textbf{71.10\%}$ & $62.71\%$ & $58.46\%$ & $54.22\%$ & $62.44\%$ \\
    BLOS-BEV Cross-Att. & $\checkmark$ & $69.75\%$ & $\textbf{65.17\%}$ & $\textbf{63.36\%}$ & $\textbf{59.81\%}$ & $\textbf{65.07\%}$ \\
    \bottomrule
  \end{tabular}
  \caption{Robustness test of \newrevise{SD} acquisition position noise on nuScenes\cite{caesar2020nuscenes} dataset. \revise{Given the GPS error, we applied a random drift of less than $10m$ and $10$° to the acquisition location of the SD map during the testing phase.} We conducted shift zero-shot tests and shift train-aug tests on all of our SD fusion methods.}
  \label{tab:robustness test}
\end{table*}

BLOS-BEV is compared against state-of-the-art (SOTA) BEV segmentation models including PON\cite{roddick2020predicting}, HDMapNet\cite{li2022hdmapnet}, LSS\cite{LSS}, and CVT\cite{zhou2022cross}. 
Tab.$\;$\ref{tab:comparation with prior sota} shows the benefits of fusing SD maps for BEV segmentation. BLOS-BEV outperforms others in both regular ($0\sim50m$) and long-range ($50\sim200m$) scenarios, demonstrating the advantages of SD maps for precise nearby and long-range segmentation. Notably, SD map fusion improves long-range segmentation by $18.65\%$ mIoU, with minimal mIoU drop-off at distances beyond the line-of-sight. This is due to the rich geometric priors in the SD map providing contextual guidance for segmentation. Our results showcase the effectiveness of fusing SD maps for accurate and robust BEV semantic segmentation at both close and long ranges.

Besides, we evaluated the performance of BLOS-BEV 
\newrevise{by employing different BEV architectures, such as HDMapNet\cite{li2022hdmapnet} and the LSS\cite{LSS}, as the BEV backbone.}
\newrevise{The experimental results in Tab.$\;$\ref{tab:comparation with prior sota} demonstrate that fusing the SD features substantially improves their performance. These findings validate the widespread effectiveness of our proposed approach to integrate SD features. }

\begin{figure}[t]
	\centering
	\includegraphics[width=0.45\textwidth]{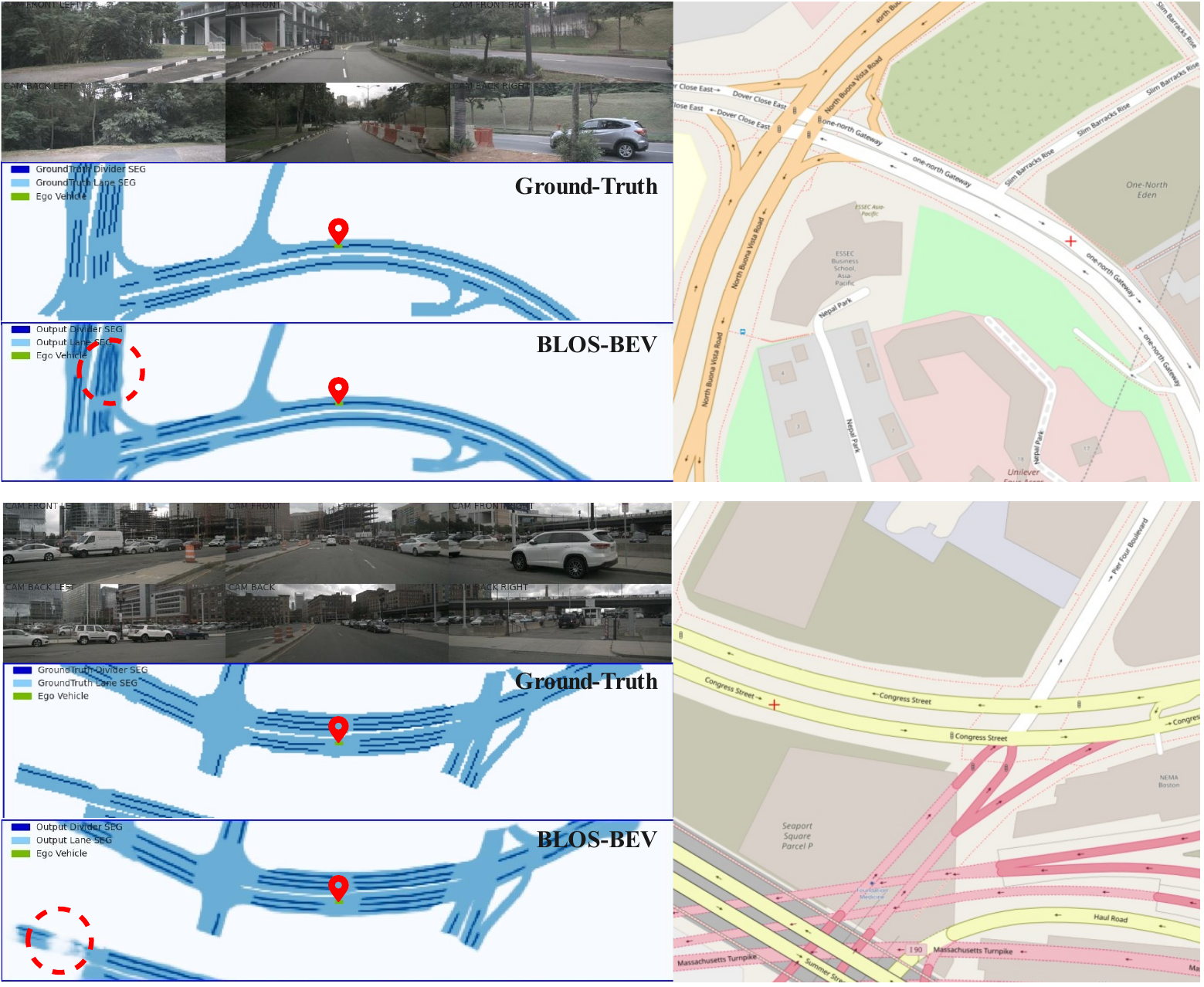}
	\caption{Extended-range BEV segmentation results from BLOS-BEV on nuScenes dataset. 
BLOS-BEV accurately labels semantic features in both close and far distances, yet some segmentation areas show slight ambiguity at long distances, marked with red dashed circles.
 }
	\label{SOLO}
\end{figure}

To visually compare the BEV segmentation results of different methods more intuitively, we present the comparative results of a scene in nuScenes in Fig. \ref{BEV comparison}.
It can be observed that without the prior information from SD maps, the segmentation effectiveness of BEV rapidly deteriorates with increasing distance. In contrast, BLOS-BEV, benefiting from the map priors, maintains robust segmentation performance even in distant predictions.
Additional generalization results are presented in Fig. \ref{SOLO}, including scenes with large curvature bends.
Such situations especially benefit from the expanded visibility of our BLOS-BEV model, substantially augmenting safety by granting extended time and space for the autonomous driving system to proactively react.

\subsection{Exploring Fusion Methods of SD Map}

We explored three different SD fusion methods with results presented in Tab.$\;$\ref{tab:exploration on fusion method of sd map}.
Notably, in the cross-attention experiment, we employed a single cross-attention layer to maintain equivalent computational complexity, ensuring a fair comparison.
Experiments on nuScenes \cite{caesar2020nuscenes} show channel-wise concatenation achieved the best performance. 
All fusion techniques provided significant gains over methods without SD maps, confirming the benefits of incorporating SD maps. 
The minor gaps between methods suggest the model can effectively leverage SD maps, regardless of the detailed fusion architecture. 

Considering the SD map has strong prior features, we also designed an experiment only using the SD map to predict BEV segmentation. The result in Tab.$\;$\ref{tab:exploration on fusion method of sd map} shows that only using the SD feature can predict an accurate road surface, but its performance is limited when predicting road boundaries, which require finer geometry. 
We conclude that the SD map provides a robust road skeleton prior, offering coarse-grained structural information vital for sensor perception. By fusing BEV and SD branches, our network achieves comprehensive and accurate environmental perception, leveraging the strengths of both.

\subsection{Performance on Argoverse Dataset}
To evaluate the effectiveness of our approach, we also conducted experiments on the Argoverse\cite{Argoverse} V1 dataset. 
Due to differences in annotation rules and categories between the Argoverse dataset and nuScenes, we cannot directly evaluate the generalization performance of nuScenes-pretrained models on Argoverse. 
Tab.$\;$\ref{tab:Argoverse_BLOS} shows the semantic segmentation results of different methods in different ranges. 
It is evident that fusing the SD map and the BEV features leads to significant improvement over the LSS baseline. 
Among the fusion methods, the cross-attention fusion mechanism achieves the best performance in all ranges.  
Moreover, for the long-range ($150\sim200m$), the cross-attention fusion of the SD map boosts the mIoU from $34.8\%$ to $60.8\%$, which is a remarkable improvement.  
Tab.$\;$\ref{tab:Argoverse_BLOS} also shows the overall segmentation results for different categories in the BLOS range. 
These results demonstrate that our method achieves excellent generalization performance across various datasets, highlighting its superior adaptability and effectiveness.

\subsection{Robustness of Position Noise}

Considering the real conditions in autonomous driving tasks, such as GPS noise, we are unable to get high accurate localization all the time.
There exists a location error when we obtain the SD map. 
This creates a gap between the BEV environment seen by the camera and the SD map, which affects segmentation performance. 
Therefore, we first conducted a zero-shot evaluation of SD location drift on BLOS-BEV. 
Next, we conducted data augmentation to fine-tune our model, specifically addressing positional noise.
We applied random position noises ($\leq10m$ and $10^\circ$) during testing and data augmentation ($\leq10m$ and $10^\circ$) during training. 
Our experiments (Tab.$\;$\ref{tab:robustness test}) show that position noise significantly impacts segmentation performance, but augmenting inputs with noise during training restores performance to the SOTA level. 
Notably, cross-attention fusion exhibits greater robustness, as it compares similar data in the database for each query, making it less affected by noisy data indices.

\section{Conclusion} 
\label{sec:conclusion}

We propose BLOS-BEV, a pioneering approach that fuses SD maps with visual perception to achieve $200m$ beyond line-of-sight BEV scene segmentation, significantly expanding the perceptible range. Our method leverages spatial context from geospatial priors to hallucinate representations of occluded regions, enabling more anticipatory and safer trajectory planning. 
Through extensive experiments and comparisons on nuScenes and Argoverse datasets, we demonstrate that BLOS-BEV achieves SOTA BEV segmentation performance at both close and long ranges.

\textbf{Limitations and Future Work.}
Currently, our work is still susceptible to the impacts of localization errors, map inaccuracies, or outdated maps, which can compromise the semantic segmentation performance for far-range BEV.
Future work will focus on exploring advanced map fusion techniques, such as incorporating HD maps where available, addressing alignment issues caused by localization errors, and investigating diffusion models to enhance BLOS effects.


\bibliography{reference.bib}

\end{document}